\documentclass{ceurart}

\usepackage{amssymb}

\usepackage{subcaption}

\usepackage{tablefootnote}

\usepackage[ruled,vlined]{algorithm2e}
\SetAlFnt{\footnotesize}

\hypersetup{breaklinks=true}
\begin{document}

\copyrightyear{2021}
\copyrightclause{Copyright for this paper by its authors.
  Use permitted under Creative Commons License Attribution 4.0
  International (CC BY 4.0).}

\conference{ALGORITHMS \& THEORIES FOR THE ANALYSIS OF EVENT DATA 2022}

\title{Enhancing Stochastic Petri Net-based Remaining Time Prediction using k-Nearest Neighbors}

\author[1]{Jarne Vandenabeele}[
]
\author[1]{Gilles Vermaut}[
]

\address[1]{Co-first authors}
\address[]{\textbf{Research Center for Information Systems Engineering (LIRIS), KU Leuven, Leuven, Belgium}}

\author[]{Jari Peeperkorn}[%
email=jari.peeperkorn@kuleuven.be
]

\author[]{Jochen {De Weerdt}}[%
email=jochen.deweerdt@kuleuven.be,
]

\begin{abstract}
  Reliable remaining time prediction of ongoing business processes is a highly relevant topic. One example is order delivery, a key competitive factor in e.g. retailing as it is a main driver of customer satisfaction. For realising timely delivery, an accurate prediction of the remaining time of the delivery process is crucial. Within the field of process mining, a wide variety of remaining time prediction techniques have already been proposed. In this work, we extend remaining time prediction based on stochastic Petri nets with generally distributed transitions with k-nearest neighbors. The k-nearest neighbors algorithm is performed on simple vectors storing the time passed to complete previous activities. By only taking a subset of instances, a more representative and stable stochastic Petri Net is obtained, leading to more accurate time predictions. We discuss the technique and its basic implementation in Python and use different real world data sets to evaluate the predictive power of our extension. These experiments show clear advantages in combining both techniques with regard to predictive power.
\end{abstract}


\maketitle

\section{Introduction}
The application of Process-Aware Information Systems (PAIS)s, such as ERP, BPMS and CRM systems to support business processes is increasing~\cite{Polato2018}. These systems record information of the process execution and possibly the individual events of that process. Drawing insights and conclusions from these data is already possible using various process mining techniques categorized into process discovery, conformance checking, and extension~\cite{VanDerAalst2012}. Within the field of process mining, predictive process monitoring concerns itself with predictive techniques applied to process data. Predominantly, three types of predictions are considered as useful: next activity, outcome, and remaining time~\cite{Marquez-Chamorro2018}. In this work we present a novel technique that combines a data-driven selection of candidate traces with building and simulating stochastic Petri nets, to predict the remaining time of running process instances. It builds upon the technique described by Rogge-Solti and Weske in~\cite{Rogge-Solti2013,Rogge-Solti2015}. This technique, referred to as \emph{generally distributed transition stochastic Petri net (GDT\_SPN)}, tailors stochastic Petri nets to incorporate the time passed since the previous completed event. We show that by combining the flexibility of GDT\_SPNs with a simple, yet effective, candidate selection approach, we can improve upon the accuracy of the predictions, as demonstrated experimentally on four real-life datasets. Our approach can be summarized as follows:
\begin{itemize}
    \item Use the K-nearest-neighbor algorithm to identify the $k$ instances most similar to the current prefix of this instance. The algorithm uses vectors based on the timestamps of previous events, while also taking into account the activity types.
    \item We use these $k$ traces to discover a (new) Petri net using the Inductive Miner~\cite{inductive} and by performing a simulation a stochastic map is obtained which complements the Petri net, to eventually obtain a GDT\_SPN.
    \item A simulation of this GDT\_SPN is further used to estimate the remaining time. This is done $n$ times and the actual prediction is taken to be the average of the $n$ simulations.
\end{itemize}

The remainder of this paper is organised as follows. In Section~\ref{Related Work}, the most relevant related work is discussed. Section~\ref{Preliminaries} defines some preliminaries, before Section~\ref{Remaining Time Prediction} presents the core of our technique, i.e., how we predict the end time, and the basic implementation of it in Python. Section~\ref{Experiments} evaluates our technique by comparing the results of different experiments with a number of benchmarks. We end this paper with Section~\ref{Conclusion}, which summarises the paper, touches upon the limitations of our technique and mentions possible further enhancements.

\section{Related Work}
\label{Related Work}

There already exist multiple techniques for predicting the remaining time of an ongoing process instance. For a comprehensive overview of the most relevant methods, interested readers are referred to Verenich et al.~\cite{Verenich2019}. For the sake of conciseness this section is limited to an overview of those techniques that are either relevant for the vast majority of the methods discussed in~\cite{Verenich2019}, or those that are highly related to the framework discussed in this paper. Earlier work concerning remaining time prediction was proposed by van der Aalst et al.~\cite{VanDerAalst2011}. By applying finite state machine techniques on event logs, they learn an annotated transition system. Such a system extends a traditional transition system with predictive information by annotating measurements of time instances at each state. Two follow-up techniques are presented in~\cite{Folino2012,Folino2013}. They enhance the technique of~\cite{VanDerAalst2011} by clustering the traces of the log in advance and creating an annotated transition system for each of these clusters afterwards. At runtime, a new trace will be assigned to a cluster and the annotated transition system for that cluster will then be used to predict the remaining time of that trace. A similar multi-stage approach is presented in this paper. The application of clustering to predictive business process monitoring has also been studied in other papers, e.g.,~\cite{Francescomarino2019}. 

Polato et al.~\cite{Polato2014,Polato2018} present three approaches, all of them using Support Vector Regressors (SVR), to predict the remaining time starting from a certain state. The first two are based on regression techniques with or without control-flow information, where the event log that serves as input is initially transformed in order to be suitable for the $\varepsilon$-SVR algorithm. The last approach is again mainly based on the idea of annotated transition systems as mentioned above~\cite{VanDerAalst2011}. It enriches each state with a Naive Bayes classifier and each transition with a Support Vector Regressor, yielding a Data-Aware Transition System (DATS).

The above mentioned techniques are often based on support vector machines and/or regression. These are, however, not the only machine learning techniques used to create predictive models. For instance, regression trees~\cite{DELEONI2016235}, XGBoost~\cite{Senderovich2017}, or random forests~\cite{Spoel2013} have already been applied for remaining time prediction. The increasing interest and latest achievements in deep learning techniques have led to a rise in predictive models using recurrent neural networks, convolutional neural networks, generative adversarial nets and, more recently, transformer nets, together with multistage approaches~\cite{AjmoneMarsan1984,Camargo2019,EVERMANN2017, tax2017, Mehdiyev_2017, Lin_2019, Pasquadibisceglie_2019, Taymouri_2020, Bukhsh_2021}. While these works show promising results, the presented models are so-called \textit{black box}, i.e. their results are hard or even impossible to interpret. This limits their use in applications where the explainable aspect is key, e.g. root-cause analysis.

We particularly highlight the work of Rogge-Solti and Weske~\cite{Rogge-Solti2013,Rogge-Solti2015}, as their approach has been the main inspiration for the technique presented in this paper. Their GDT\_SPN-formalism based technique differs from other approaches as it takes into account the time passed since the last observed event, while the other approaches only update predictions upon arrival of finished events. To achieve this, stochastic Petri nets are equipped with generally distributed transitions, in which distributions reflect the duration of the corresponding activities in the real world and can be different from the exponential distribution, i.e., the Markovian property is not enforced. The latter distributions may then be updated, based on the time passed since the last observed event, to achieve more accurate predictions. These models are not \textit{black box}, and can thus be used to explain multiple aspects influencing the execution time of a certain instance. 

\section{Preliminaries}
\label{Preliminaries}

In this chapter, we describe the preliminary concepts on which our novel prediction technique is based. As mentioned above, the goal of this technique is to predict the remaining time of ongoing cases. This entails that the predictive model can only use information of the current case up until this point in time, supplemented with information from past cases. The assumption is made that traces contain timestamps for each occurring event. Those timestamps indicate at least the end, and in some cases the start, of the activities represented in the trace. Each event in the event log should contain a \emph{trace ID}, indicating to which process execution it belongs, together with an activity type. %This is all the framework needs as it solely relies on the timestamps related to the events of each instance to predict the remaining time. 
Our technique is an extension of the work of Rogge-Solti and Weske, who introduce the use of stochastic Petri nets with generally distributed transitions, so-called GDT\_SPN models, to make predictions~\cite{Rogge-Solti2013,Rogge-Solti2015}. In this technique, a Petri net is enriched with statistical timing data for each event, which allows end-users to make time predictions. The definition of both a Petri net and a GDT\_SPN as presented by Rogge-Solti and Weske is given below~\cite{Rogge-Solti2015}: 

\vskip 0.1in
\textbf{Definition 1 (Petri Net)} A Petri net is a tuple PN = (P, Tr, F, M$_{0}$) where:
\begin{itemize}
  \itemsep-0.0em 
  \item P is the set of places.
  \item Tr is the set of transitions.
  \item F $\subseteq$ (P \texttimes Tr ) $\cup$ (Tr \texttimes P) is the flow relation.
  \item M$_{0} \in P \rightarrow \mathbb N_{0}$ is the initial marking.
\end{itemize}

The Petri net models used by~\cite{Rogge-Solti2015}, and hence as well in our technique, are restricted to sound workflow nets. 

\bigskip
\textbf{Definition 2 (Generally Distributed Transition Stochastic Petri Net)}
A GDT\_SPN is a seven-tuple: GDT\_SPN = (P, Tr, $\mathcal{P}, \mathcal{W}$, F, M$_{0}, \mathcal{D}$), where (P, Tr, F, M$_{0}$) is the basic underlying Petri net as specified above. Additionally:
\begin{itemize}
  \itemsep-0.0em
  \item The set of transitions Tr is split into immediate transitions Tr$_{i}$ and timed transitions Tr$_{t}$.
  \item $\mathcal{P}$ : Tr $\rightarrow \mathbb N_{0}$ is the assignment of priorities to the different transitions, where $\forall \tau \in$ Tr$_{i}$: $\mathcal{P}(\tau) \geq$ 1 and $\forall \tau \in$ Tr$_{t}$: $\mathcal{P}(\tau)$ = 0.
  \item $\mathcal{W}$ : Tr$_{i} \rightarrow \mathbb{R}^{+}$ assigns probabilistic weights to the immediate transitions Tr$_{i}$.
  \item $\mathcal{D}$ : Tr$_{t} \rightarrow$ D is the assignment of probability distribution functions D to timed transitions Tr$_{t}$, reflecting the durations of the corresponding activities.
\end{itemize}

These probability distribution functions D do not need to be exponentially distributed, but can also be normal distributions, uniform distributions, etc. and hence do not necessarily have the Markovian property, i.e., memorylessness. This is an important factor as the absence of this property is used to update the distribution functions based on the passed time. This is the key difference with the more known generalised stochastic Petri nets (GSPN), as presented by Marsan et al.~\cite{AjmoneMarsan1984}

Rogge-Solti and Weske~\cite{Rogge-Solti2013,Rogge-Solti2015} exploit the idea of memorylessness to update the density function of the original distribution towards a new density function of a \textit{truncated} distribution. The intuition behind this is that when some time has passed, the probability increases that the activity will be completed in the nearer future as some work might already have been done. This is in contrast with the above mentioned Markovian property. This is done by using the elapsed time since the last event. Let F$_\delta(t)$ be the  duration distribution function of that activity. By differentiating F$_\delta(t)$ you can obtain a density function f$_\delta(t)$. We then use $t_{0}$, the current time since enabling the transition, to truncate the distribution as follows~\cite{Rogge-Solti2015}:
\[ f_{\delta}(t|t \geq t_{0}) = \left\{
    \begin{array}{lr}
        0 &  t < t_0,  F_{\delta}(t_0) < 1 \\
        \left.\frac{f_\delta(t)} {1 - F_\delta(t_0)} \right. &  t \geq t_0,  F_\delta(t_0) < 1 \\
        f_{\delta_{Dirac}}(t - t_0) &  F_\delta(t_0) = 1
    \end{array}
    \right.
\]

The part of the density function above (or in this case more correctly \textit{after}) $t_0$ is rescaled in such a way that it integrates to 1. For the case where F$_\delta(t_0) = 1$, i.e. when the current time has progressed further than the density functions supports, we use the Dirac delta function $f_{\delta_{Dirac}}$ (a function whose value is $0$ everywhere except at one peak, and whose full integral is equal to $1$), with a peak at $t_0$. This happens when the current event is taking longer to complete than the events in the training log corresponding to the same activity type. The basic idea is that the predictions are only based on those cases for whom the corresponding activity would not have been already completed at the current time $t_0$. For a more extensive explanation, together with the effects of this truncation of different types of distributions, interested readers are referred to~\cite{Rogge-Solti2015}. In addition, \textit{n} is the number of simulations performed by the GDT\_SPN, for a given sample case. The eventual prediction is equal to the mean duration. As will be explained in Section~\ref{Remaining Time Prediction}, our proposed prediction technique combines the algorithm as described by~\cite{Rogge-Solti2013,Rogge-Solti2015} with the basic idea of the \textit{k}NN algorithm. We have adopted the notion of finding the \textit{k} most representative training instances for the to-be-predicted instance. These \textit{k} representative cases found during \textit{k}NN are used to build a predictive GDT\_SPN model that can then be used to make a prediction.

\section{Remaining Time Prediction using GDT\_SPN\_\textit{k}NN}
\label{Remaining Time Prediction}
In this Section, we introduce generally distributed transition stochastic Petri net with \textit{k}NN-based candidate selection, referred to as GDT\_SPN\_\textit{k}NN. The algorithm depends on the number of neighbors taken into account, hyperparameter \textit{k}. In order to make predictions, the algorithm further requires a log file containing complete traces of the business process (the \textit{training log}), the time passed since the start of the case \textit{t$_0$}, and a partial trace \textit{T}, containing events with timestamps until \textit{t$_0$}. The key extension of GDT\_SPN\_\textit{k}NN with respect to the original GDT\_SPN algorithm, resides in the fact that we avoid using the whole training log as an input to construct the GDT\_SPN model. In contrast, in GDT\_SPN\_\textit{k}NN, only the \textit{k} complete traces that are most similar to the to-be-predicted partial trace \textit{T} are taken into account. The main motivation for selecting this subset is that, under the condition that the right features are selected, only looking at the \textit{k} most similar traces will yield a predictive model that embodies less variation in the generalised density functions, likely leading to a final prediction closer to the real value. If tuned well, selecting only the \textit{k} nearest instances thus omits irrelevant cases and outliers and yields a more representative and stable GDT\_SPN model for the given partial trace. %It is assumed that time until detection of completed events is almost zero. \\

The choice of the hyperparameter \textit{k} to decide the number of nearest neighbors is not obvious.  When \textit{k} is taken too small, the event distributions built by the \textit{k} neighbors may be inaccurate or have a high variance, which leads to volatile and most of the time worse results. When \textit{k} is taken too large, the effect of looking at only the most similar traces may be minimal or even non-existent, given that the larger \textit{k}, the more the outcomes will look like those of the original technique. By taking \textit{k} too large, the variance may increase as well, as multiple trace variants are taken into account for building the model, which leads to a more complex Petri net. These different trace variants may also have a different underlying distribution to estimate the duration of the activities. Another drawback of setting \textit{k} too large is that the computation time increases, as more traces need to be replayed to construct the Petri net and distributions. Nonetheless, despite the fact that hyperparameter \textit{k} is key, it is computationally demanding to tune it. Based on initial explorations, we found that a value of 100 provided to be a good trade-off value. While in a practical application, it would be certainly beneficial to tune \textit{k} carefully, in this paper, mainly due to a lack of time, we work with this fixed value of 100. In further research, we would like to investigate dedicated strategies to tune \textit{k}, beyond an exhaustive search. 

%By varying the value of \textit{k}, we found that for the experiments on the data sets used in this paper, 100 provided to be a good trade-off value. Due to time constraints, we do not further elaborate on this. However, it should be noted that this value depends on the nature of the training set, the underlying process and the amount of variation displayed by it, and that therefore one should take care while setting this parameter. 

For applying \textit{k} nearest neighbors, a proper featurization should be applied. While a conventional feature engineering would typically consider trace and event attributes, we propose a method that only relies on time information. More specifically, on a per event basis, we take into account the total time from the start of each trace \textit{t$_0$} until the occurrence of that individual event. This total duration between the start of the \textit{trace} and the occurrence of that \textit{event} will be referred to as the \textit{time-to-occurrence} of an event in the remainder of this paper. In case of the presence of loops, the \textit{time-to-occurrence} is determined based on the last repeated activity observed in a trace. 
The rationale behind using \textit{time-to-occurrence} is that there is a likely correlation between traces that have a similar time-to-occurrence for events corresponding to the same activity type and hence, it might thus be valuable to sample them in order to make better remaining time predictions. This feature engineering is also more generalisable compared to matching neighbors based on attributes, as not all log files contain event and/or trace attributes. However, for certain processes, it might well be that taking into account other trace or event features is beneficial. Our algorithm easily allows to incorporate this, when e.g. expert knowledge suggests correlations between these features and the remaining time. When we eventually build the GDT\_SPN using the \textit{k}NN, we still do \textit{n} different simulations for which we take the average values as our eventual prediction. In this work, the value of $n = 500$ is chosen, adopted from~\cite{Rogge-Solti2015}, as taking this high enough will make the predictions more robust and less volatile.

\SetKwProg{Fn}{Function}{}{end}

%\bigskip
\begin{algorithm}[ht]
\SetAlgoLined
\KwResult{Times\_to\_occurrence vector $\psi^*$ for a trace $\psi$}
$\psi$ = Trace \;
Voc = [all activity types in training data]\;

\Fn(){Trace\_times\_to\_occurrence($\psi$, Voc)}{
$\psi_{start}$ = start\_time($\psi$)\;
$\psi^* = [\psi^*_1, \psi^*_2, \dots, \psi^*_D]$ with $D = |\text{Voc}|$ \;
 \For{$i \in \{1, \dots, D\}$}{
 \eIf{$\text{Voc}_i \in \text{activities}(\psi)$}{
 event = event corresponding to the last occurring event of activity type $\text{Voc}_i$\;
 endtime = timestamp(event)\;
 $\psi^*_i$ = endtime - $\psi_{start}$\;}
 {$\psi^*_i = -1$\;}
 }
 \Return{$\psi^*$}
}

 \Fn(){activities($\psi$)}{
 Intitialize: $\alpha = [\alpha_1, \alpha_2, \dots, \alpha_L]$ with $L = |\psi|$ \;
 \For{$i\in \{1, \dots, |\psi|\}$}{
 $\alpha_i$ = Activity type of event $\psi_i$
 }
 \Return{$\alpha$}
 }
 \caption{Feature Construction of the times\_to\_occurrence vector for a trace }
 \label{feature_construction}
\setlength{\abovecaptionskip}{-8pt}
\setlength{\belowcaptionskip}{-8pt}
\end{algorithm}

\begin{comment}
%\bigskip
\begin{algorithm}[ht]
\SetAlgoLined
\KwResult{Pre-processed data}
 all\_traces\_times\_to\_occurrence = list()\;
 activity\_vocabulary = [all activity types in training data]\;
 activities(trace) = [all activity types in that trace]\;
 \For{$\psi$ $\in$ training data}{
  %$\psi$ = training\_trace\;
  $\psi_{start}$ = start\_time($\psi$)\;
  trace\_times\_to\_occurrence = list()\;
  \For{a $\in$ activity\_vocabulary}{
  \eIf{a in activities($\psi$)}{
   event = event corresponding to the last occurring event of activity type a\;
   endtime = timestamp(event)\;
   trace\_times\_to\_occurrence.add(endtime - $\psi_{start}$)\;
   }{
   trace\_times\_to\_occurrence.add(-1)\;
   }}
    all\_traces\_times\_to\_occurrence.add(trace\_times\_to\_occurrence)\;
   }
 \caption{Feature Construction of the Training Log}
 \label{feature_construction}
\setlength{\abovecaptionskip}{-8pt}
\setlength{\belowcaptionskip}{-8pt}
\end{algorithm}
\end{comment}

\begin{comment}
\begin{equation*}
\begin{split}
    \forall \text{ traces } \psi \in \text{Training Log}:& \psi^* = \underbrace{(\psi^*_i, \psi^*_i, \dots, \psi^*_i)}_\text{|Voc|} = \underbrace{(-1, -1, \dots, -1)}_\text{|Voc|} \\
    &\forall i \in \{1, \dots, |\text{Voc}|\}: \psi^*_i = \begin{cases}
  Timestamp & \text{if } v \in L
\\
  0 & \text{if } v \not\in L
\end{cases}
\end{split}
\end{equation*}
\end{comment}

The feature construction needed to predict the remaining time of a (partial) test trace \textit{T} can be seen in Algorithm~\ref{feature_construction}. We define an activity vocabulary \textit{Voc} containing all possible activity types present in the process. Each trace $\psi$ of the training log is formatted as follows. For each activity present in the activity vocabulary of the event log, the algorithm checks whether an event occurs in $\psi$ corresponding to that activity type. If it does, then the difference between the time of this event in $\psi$ and the beginning of this trace is calculated and stored in a formatted vector $\psi^*$ corresponding to this trace $\psi$. If it does not, a negative value is assigned. We call this value the \textit{time-to-occurrence} of that activity type ($\psi^*_i$ in the algorithm above). For the to-be-predicted trace \textit{T} a similar formatted vector is constructed. If a certain activity type is present in the trace multiple times, as is the case for e.g. loops, we only consider the last occurring event. We choose to take the last occurrence, because if a certain activity has to be performed multiple times (in one execution), we assume most often the final completion time of that activity (i.e. the timestamp of the last occurrence) provides the most useful information. However, this choice can be easily altered if needed.

The result of this procedure is that for each trace, we obtain a formatted counterpart in the form of a vector with a fixed length. The length corresponds to the number of activities in the business process, i.e., all distinct activities observed in the training data. Each entry in this vector corresponds to a fixed activity. The entry is positive if an event corresponding with the completion of the activity appears in the trace, and is negative otherwise. For the remainder of this explanation, we call the formatted version of the to-be-predicted trace \textit{$T^*$}. Only the distance between relevant events is measured. These are the events that appear in the partial test trace \textit{T} and consequently have a positive value in \textit{$T^*$}. We want traces from the training log that follow the same path, i.e. having the same executed activities, as the partial trace to have a higher impact, since these traces are more likely to be similar to the partial trace in question. Accordingly, a penalty, in the form of a maximal distance vector, is induced on those that do not follow the same route up to the point of prediction as the partial test trace. This means that the assigned formatted vector for such a trace will contain a large constant for each event, i.e., the maximum time-to-occurrence seen in the training set, making this trace very far off when selecting the nearest neighbors. A min-max normalisation is applied on the formatted versions of the training traces and on \textit{$T^*$}, which gives all events equal influence in calculating the distance. This makes them suitable for the \textit{k}NN algorithm. If \textit{k} is higher than the number of traces present in the training set that follow the same route, the algorithm randomly selects the other traces up to \textit{k}. In a future extension, this might be replaced by taking prefixes closest to the control-flow of prefix $T$, by some metric. We use the Euclidean distance between the formatted vectors as a distance function in the \textit{k}NN algorithm. In summary the executed procedure has the following characteristics: 
\begin{itemize}
    \itemsep0em
    \item Only the distance between relevant events is measured. These are the events that appear in the partial test trace \textit{T} and consequently have a positive value in \textit{$T^*$}. The formatted versions of the training traces can have positive values for other events as well, but these are not taken into consideration when selecting the nearest neighbors, as only the events present in \textit{T} are used to calculate the distances. 
    \item We want traces from the training log that share all relevant events with the partial trace to have a higher impact, since these traces are more likely to be similar to the partial trace in question. That is why a penalty is induced on those that do not follow the same route up to the point of prediction as the partial test trace. It should be noted that the rare situation may occur where the number of \textit{k}NN is higher than the number of traces present in the training set that follow the same route. If this situation happens, the algorithm will randomly pick the other traces up to \textit{k}. This is another motivation for proper parameter tuning and to keep the parameter \textit{k} small enough.
    \item Because of the normalisation of the vectors, one should remark that all completed events have an equal influence while calculating the distance. In the case an outlier is present, the distance between normal traces will be rather small. 
    \item An important consequence of taking the nearest neighbors to build our model is that we allow the process to be dynamic. The occurrence of a new trace variant in the business process will be fully taken into account in the model as soon as \textit{k} training examples of this trace variant are recorded. %However, a likely poor prediction is already possible after observing only a single completed instance of that trace variant. When dealing with highly dynamic models, this is another argument for keeping \textit{k} reasonably low.
    \end{itemize}

When the nearest neighbors are found, the full training traces corresponding to these neighbors are used to create a Petri net using Inductive Miner~\cite{inductive}, which guarantees to produce sound and fitting models and hence alleviating some of the shortcomings of the $\alpha$-miner~\cite{Leemans2013}. As mentioned in Section~\ref{Preliminaries}, this soundness is a necessity for the Rogge-Solti and Weske algorithm~\cite{Rogge-Solti2013,Rogge-Solti2015}. Additionally, fitting models make it possible to replay the partial trace \textit{T}, which is necessary for the algorithm to make a prediction~\cite{Buijs2014}. Given this Petri net and \textit{k} training traces, simulation can be performed to obtain a stochastic map to complement the Petri net. A stochastic map projects every activity on a probability distribution function. In this way, we obtain a GDT\_SPN that can be used for prediction using the original simulation of Rogge-Solti and Weske. Here we have to set another hyperparameter $n$, which is equal to th number of simulations we perform for the partial trace. The actual prediction is then taken as the average of all these simulations. For the rest of this work we use $n$ equal to 500, which should be high enough to average out outliers in the simulation.

We have translated the implementation of Rogge-Solti and Weske~\cite{Rogge-Solti2013,Rogge-Solti2015}, that was available in the open-source program ProM, to a standalone implementation in Python compatible with the \textit{pm4py} framework~\cite{Berti2019}. It does not yet fully cover all features that are available in ProM, but it covers the core algorithm and achieves similar results. Our standalone implementation, including the addition of the \textit{k}NN algorithm, can be found on Github\footnote{https://github.com/JarneVDB/BP-Time-Prediction-using-KNN}. For the \textit{k}NN algorithm we use the implementation in Scikit-Learn~\cite{scikit-learn}.

Our approach is agnostic to the specific candidate selection-technique, given that  \textit{k}NN could be replaced by another technique. For instance, this might be an eager clustering technique where for each cluster a GDT\_SPN model can be constructed during the training phase. Similar approaches have been explored in~\cite{Folino2012,Folino2013,Francescomarino2019}. One advantage of taking such an approach is that, as opposed to the lazy learner \textit{k}NN, this would yield a better performance in the deployment phase. However, there are also reasons why using \textit{k}NN could result in more accurate predictions. Eager clustering techniques often yield large sized clusters in addition to some smaller clusters, which takes away the power of combining smart candidate selection with the approach of~\cite{Rogge-Solti2013,Rogge-Solti2015}. Additionally, using \textit{k}NN allows the process to be dynamic, as changes can be picked up quickly in the resulting GDT\_SPN model, leading to more accurate predictions. Depending on the importance of stressing either performance or accuracy and flexibility, one can choose an appropriate kind of learner when adapting our approach.

\section{Experimental Evaluation}
\label{Experiments}

\subsection{Experimental setup}
\begin{figure*}[ht]
\label{Evaluation setup}
\begin{center}
\includegraphics[width=\linewidth]{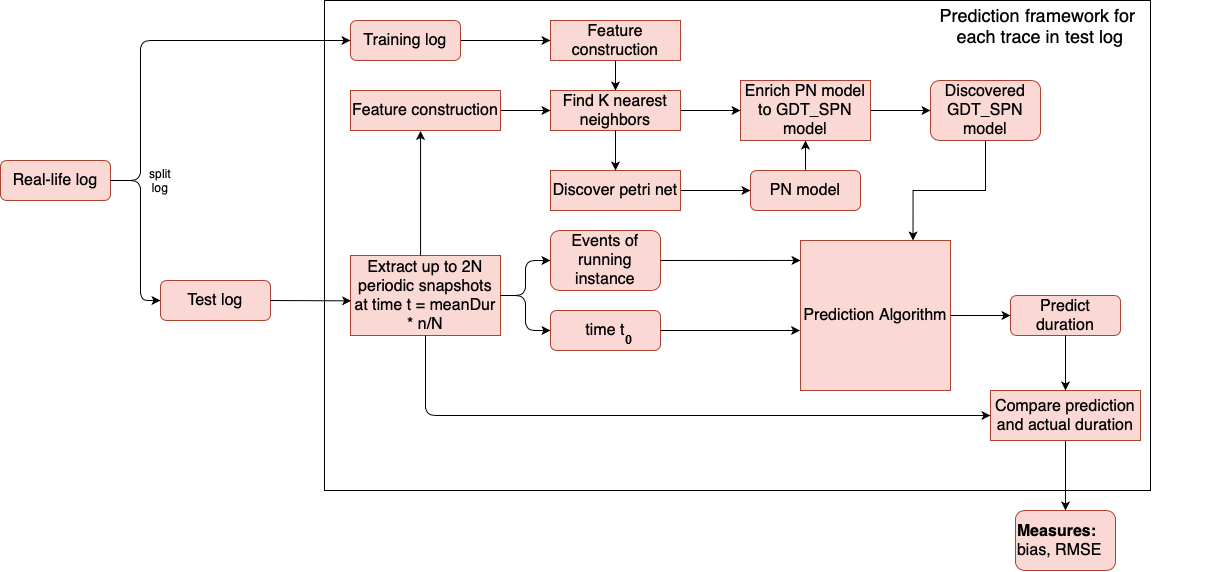}
\end{center}
\hfill
\setlength{\abovecaptionskip}{-8pt}
\setlength{\belowcaptionskip}{-8pt}
\caption {Evaluation setup}
\vspace{-0.5cm}
\label{fig:evaluation}
\end{figure*}

In this section, we will discuss the results of our approach on four real-life datasets. A schematic overview of the setup is given in Figure~\ref{fig:evaluation}. 

Each real-life dataset is split into a training and a test log. In order to compromise between a large enough test set and reasonable computing time, all experiments use a test log with approximately 500 traces, independent of the amount of traces in the training log. The size of the training log is always significantly larger than the size of the test log, ranging from 2500 to 8000 traces, depending on the size of the original dataset from which we took a subset. The train and test log split was done out-of-time, in order to avoid possible data leakage. The training log used to obtain the \textit{k} neighbors only contains finished traces upon to that point in time. All following steps are repeated \textit{x} times, with \textit{x} corresponding to the number of traces in the test log. For each trace in the test log, at most $2N$ periodic prediction iterations will be performed, where $N$ is a hyperparameter that influences the number of prediction evaluation iterations. Corresponding to the methodology of~\cite{Rogge-Solti2015}, the Nth prediction will be performed \textit{meanDur} after the start of the case, where \textit{meanDur} corresponds to the average case duration of the training log. For all experiments, we set N equal to 20, leading to 40 prediction iterations. Each iteration thus refers to a point in time after the start of the trace, where we predict the remaining item of this execution. 2N can thus be regarded as the number of different points in time where we decide to predict the remaining lead time. Each time we do perform a prediction we provide the algorithm with the current traces up to $t_0$ (corresponding to this point in time). We call the specific point in time we are calculating the remaining times of all (still) ongoing cases, the prediction iteration. Hence, at most 40 moments of prediction are simulated for each trace, each corresponding to different stages in the execution of the process instance. When the execution has finished, at some iteration, no more predictions are done for this particular trace, thus most of the cases in the test log will only influence part of the prediction iterations (as they will be finished at some point). It should be noted that when the process has a low variance, the effective number of prediction iterations for each test trace will be close to $N$. When increasing the iteration, less and less traces are used to evaluate the time prediction, making the the results more volatile and statistically less significant. 

For both the partial test trace \textit{T} and the entire training log, feature construction is executed as described in Section~\ref{Remaining Time Prediction}, yielding the corresponding vector representations with the times-to-occurrence of all known activity types. In the transformed version of the training log, the 100 nearest neighbors of the formatted test trace \textit{$T^*$} are found. The full traces corresponding to these neighbors are used to discover a Petri net, which is enriched to a GDT\_SPN model by means of stochastic information resulting from a simulation. Although the algorithm is flexible in the choice of distribution types, we decided to force a normal distribution for all non-immediate events. The choice of the most proper distribution is most likely event log (process) dependent, but the normal distribution was chosen for every process in this paper due to time constraints. A further investigation on this topic, might clarify certain (possible) issues. 

\begin{figure*}[htp]
\begin{center}
\label{Petri Net BPI 2019}
\includegraphics[scale=0.39]{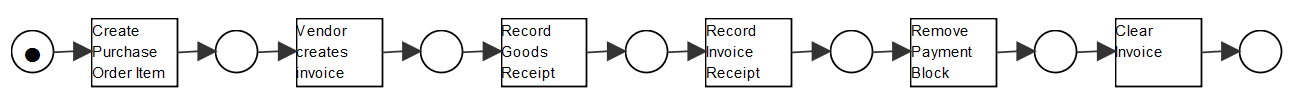}
\end{center}
\caption {Petri net of the selected control-flow variant in the BPI Challenge 2019 dataset, illustrating a purchase order business process.}
\label{fig:purchase_order}
\end{figure*}

The results of the predictions are better when forcing normal distributions, given that the effect of the model vanishes when using memoryless exponential distributions and alternative distributions such as the uniform distribution are sensitive to outliers. Moreover, we assume that the normal distribution is the distribution type that is often a good estimation for the real underlying distribution function. This GDT\_SPN model, together with the events of the running instance \textit{T} and time \textit{t$_0$}, serve as input for the prediction algorithm as described in~\cite{Rogge-Solti2013,Rogge-Solti2015}. The number of different simulations performed by each GDT\_SPN \textit{n}, of which the purpose is explained above, is set to 500 for each experiment as this averages out most of the influence of outliers. The reported evaluation metrics will be the average error and the root mean square error (RMSE). These two metrics will allow us to respectively  measure the bias and accuracy of our prediction method. Four different event logs are used in the experimentation. The first one, as a simple proof-of-concept, uses the BPI Challenge 2019 Event Log. However, instead of using the full event log, one single control-flow-variant is selected, which can be seen in Figure~\ref{fig:purchase_order}. The experiments on the other selected event logs, do use multiple different control-flow variants. These event logs are the \textit{Hospital log}, depicting the billing process in a hospital, and the BPI Challenge 2020 event logs, depicting the travel expense declaration process of a university for domestic (\textit{BPI2020d}) and international travel (\textit{BPI2020i}). An overview of summary statistics regarding the different data sets can be found in Table~\ref{table:datasummary}. 

\begin{table}[!ht]
\begin{center}
    \begin{tabular}{p{1.85cm}p{1.0cm} p{1.2cm} p{1.1cm} p{1.1cm} p{1.1cm} p{1.1cm} p{1.1cm} }
    \hline
     Datasets & Cases & Events & Event classes & Max case length & Avg. case length & Max case time & Avg. case time \\ \hline
    \textit{BPI2019\tablefootnote{\url{https://doi.org/10.4121/uuid:d06aff4b-79f0-45e6-8ec8-e19730c248f1}}} \\(1 variant) & 7,460 & 44,760 & 6 & 6 & 6 & 356.21 & 94.54\\ \hline
    \textit{Hospital\tablefootnote{\url{https://doi.org/10.4121/uuid:76c46b83-c930-4798-a1c9-4be94dfeb741}}} & 7,847 & 33,450 & 7 & 6 & 4.26 & 867.54 & 156.95\\ \hline
    \textit{BPI2020d\tablefootnote{\url{https://doi.org/10.4121/uuid:3f422315-ed9d-4882-891f-e180b5b4feb5}}} & 7,820 & 40,281 & 7 & 6 & 5.15 & 290.89 & 10.52\\ \hline
    \textit{BPI2020i \tablefootnote{\url{https://doi.org/10.4121/uuid:2bbf8f6a-fc50-48eb-aa9e-c4ea5ef7e8c5}}} & 2,361 & 23,726 & 14 & 12 & 10.05 & 463.04 & 80.17\\ \hline
    \end{tabular}
\end{center}
\setlength{\belowcaptionskip}{-8pt}
\caption{Descriptive statistics of event logs used for the training of the model. Time-related characteristics are reported in days.}
\label{table:datasummary}
\end{table}

\subsection{Experiment Results}
\begin{figure*}[ht]
\begin{center}
\includegraphics[width=\linewidth]{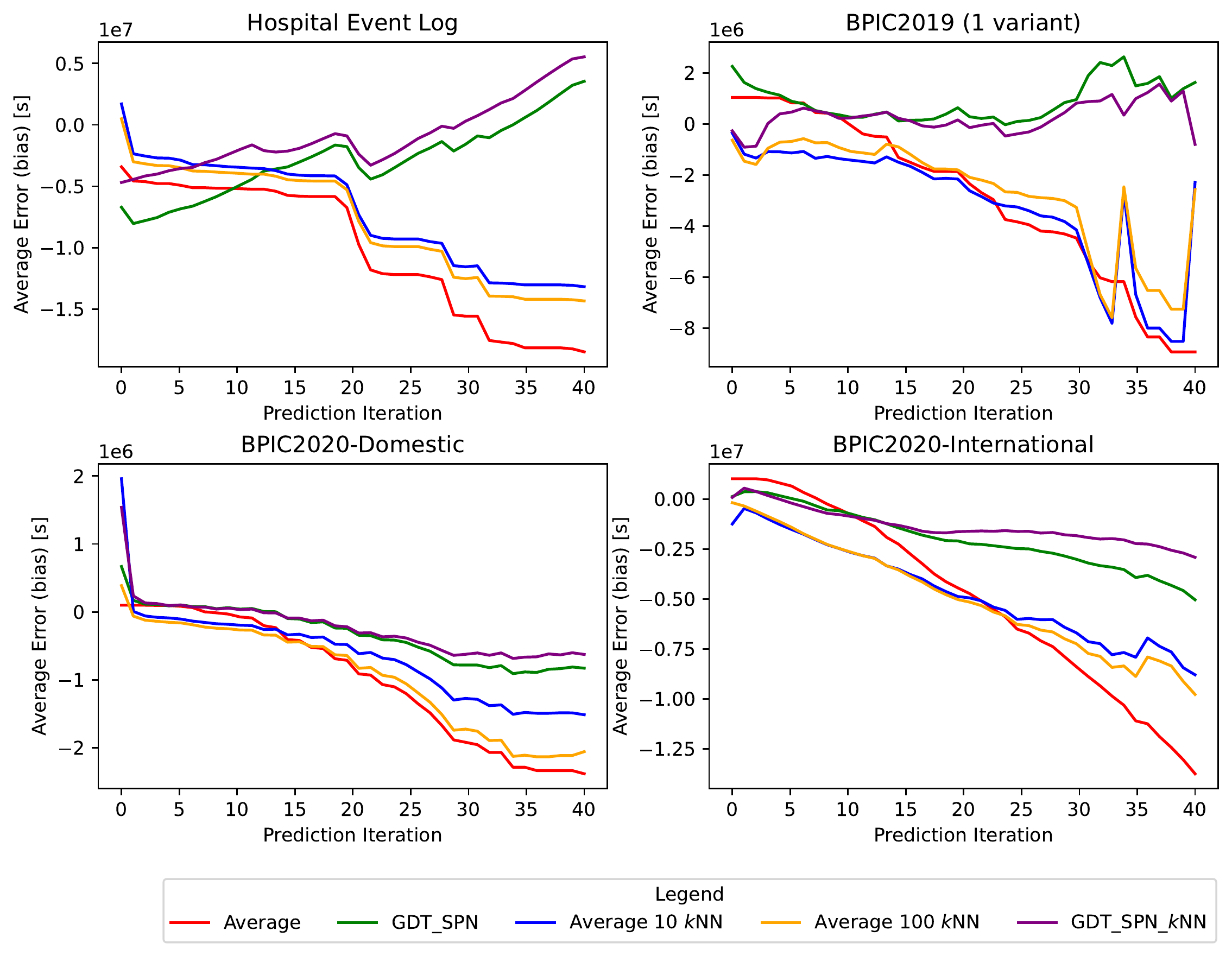}
\end{center}
\hfill
\setlength{\abovecaptionskip}{-8pt}
\setlength{\belowcaptionskip}{-8pt}
\caption{Mean errors of real-life experiments.}
\label{mean_errors}
\setlength{\abovecaptionskip}{-8pt}
\setlength{\belowcaptionskip}{-8pt}
\end{figure*}

\begin{figure*}[ht]
\begin{center}
\includegraphics[width=\linewidth]{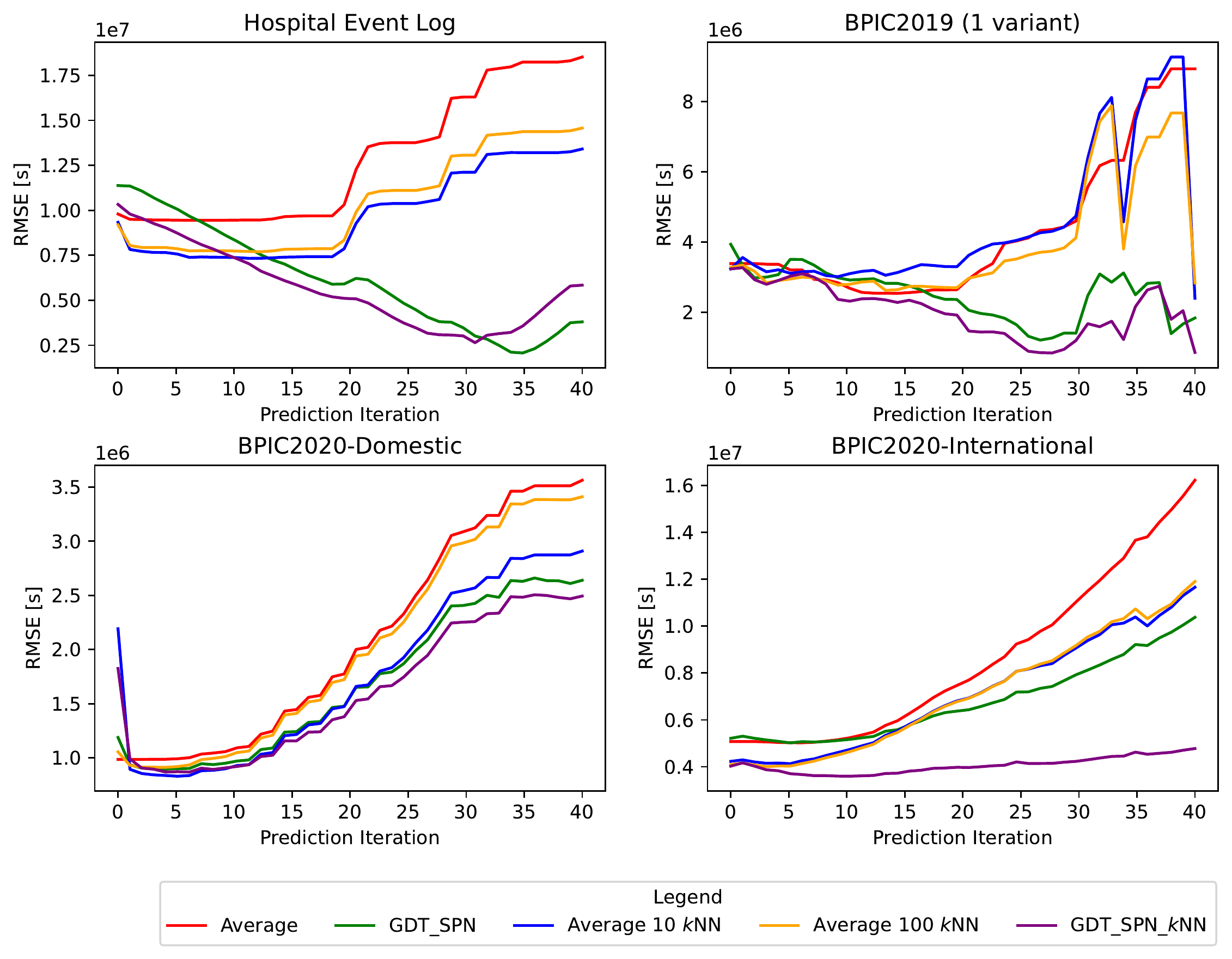}
\end{center}
\hfill
\setlength{\abovecaptionskip}{-8pt}
\setlength{\belowcaptionskip}{-8pt}
\caption{Root mean square errors of real-life experiments.}
\label{rms_errors}
\setlength{\abovecaptionskip}{-8pt}
\setlength{\belowcaptionskip}{-8pt}
\end{figure*}

GDT\_SPN\_\textit{k}NN is tested against four benchmarks: the average duration of the full training set (Average), the average duration of the ten nearest neighbors of the candidate trace's prefix (Average 10 \textit{k}NN), the average duration of the hundred nearest neighbors (Average 100 \textit{k}NN), and the algorithm as proposed by Rogge-Solti and Weske (GDT\_SPN)~\cite{Rogge-Solti2013,Rogge-Solti2015}. One should note that for later iterations, only few to-be-predicted traces remain in the test set and that the results can thus be volatile and less informative towards the very end. The outcome of these experiments is visualised in Figure~\ref{mean_errors} and Figure~\ref{rms_errors}, where the former reports the mean errors and the latter reports the root mean square errors. Note that in these figures our prediction algorithm is referred to as \textit{GDT\_SPN\_\textit{k}NN}. The x-axis represents the number of prediction iterations (as explained above), while the y-axis expresses the value of the corresponding metric in seconds. 

In Figure~\ref{mean_errors} and Figure~\ref{rms_errors} no systematic under- or overestimation is observed. The bias compared to the benchmarks becomes smaller as more information about activities is known, i.e., as the prediction iteration goes up. In the vast majority of the prediction iterations, we achieve a higher accuracy than all benchmarks. The better predictions can be explained by two main factors. First, when there exists correlation between the times-to-occurrence of the different events, the algorithm can exploit this as soon as the first time-to-occurrence becomes available. 

A second factor is that the Petri net for each cluster is more simple since it uses fewer trace variants, whereas the original GDT\_SPN constructs a Petri net out of all training traces. This has a consequence in the further simulation of the Petri net as it is more likely that the ending is fixed, while in the more complex Petri net of~\cite{Rogge-Solti2013,Rogge-Solti2015}, multiple paths can be followed towards the final marking that actually belong to other trace variants. These trace variants might contain events portraying activity types not present in the test trace and might therefore yield worse predictions. Whenever the different trace variants have some matching activities, this factor becomes particularly important as the further simulation of the complete Petri net as in~\cite{Rogge-Solti2013,Rogge-Solti2015} would be too volatile. 
It should be noted that although the results on the above data sets, on average, are showing significant improvements with regard to the stated benchmarks, there are some cases in which our approach does not yield better results. The predictions taken earlier on, at a lower iteration, show less difference between the tested methods as well. The benchmarks using simple averaging score not far off from the more elaborate GDT\_SPN based methods. Later in the traces (for a higher iteration), the advantage of incorporating the time already passed on an activity, and possibly other information concerning the activities yet to come, result in more accurate predictions.  %When both the number of paths with some matching activities and the correlation between the activities are low, our method does not yield any improvement compared to the approach of Rogge-Solti and Weske, as the algorithm does not find indicative neighbors and a similar Petri net is built as in~\cite{Rogge-Solti2013,Rogge-Solti2015}. 

\section{Conclusion and future work}
\label{Conclusion}

In this paper, we constructed an approach for predicting the remaining time of a business process based on a combination of nearest neighbor selection and the GDT\_SPN model as presented in~\cite{Rogge-Solti2013,Rogge-Solti2015}. According to the four datasets we used to validate our model, we significantly outperform our four benchmarks, including the original method~\cite{Rogge-Solti2013,Rogge-Solti2015}. The higher the correlation between the different times-to-occurrence and the more path variants with similar activities, the better our algorithm performs compared to the benchmarks. The GDT\_SPN model itself is \textit{white box}, and can therefore be used for explainability purposes. And while the \textit{k}NN selection adds some complexity to the methodology, this does not obstruct explainability, and might even improve it when the discovered Petri nets are more simple.

Multiple assumptions were made in the experiments presented in this paper, such as putting the number of neighbors $k = 100$. The impact of these choices could be investigated further. Next to this, there are still multiple ways to build further on the prediction method presented in this paper, as both the choice of distance metric, clustering method and even the choice of using a GDT\_SPN instead of something something else, is flexible and can easily be interchanged. One could develop a way to adjust the weighting of the activities by putting more weight on more relevant activities. This could potentially lead to better neighbor selection and hence better prediction. One could take into account trace and event variables while selecting nearest neighbors. A possible way of doing this is by a nested clustering approach, were one first clusters on either attributes or times-to-occurrence and afterwards reduces the number of neighbors by clustering on the other one. Furthermore, these extra attributes could be used to select the most relevant prefixes, when less than $k$ prefixes can be found whose control-flow correspond to that of the case in question. In order to improve scalability, one could create fixed clusters and assign each test trace to the cluster it is most similar to. With such an eager clustering approach instead of the \textit{k}NN algorithm, each cluster would have its own GDT\_SPN model. These models can be built upfront and can directly be used to make a prediction from the moment the test trace is assigned to the corresponding cluster. While this may increase performance, the accuracy may drop, especially when the process is dynamic. Moreover more experimentation into the impact and sensitivity of the results with different parameter values when setting up the inductive miner and different ways of matching neighbors, could provide some interesting insights. Moreover, investigating the true duration distributions might be interesting, as in this work we assumed Normal distributions. If the true distribution is not normal, learning different kinds of parametric (or even non-parametric) models might increase the prediction accuracy. 

%% Define the bibliography file to be used
\bibliography{Main.bib}

\end{document}